\documentclass[sigplan,screen]{acmart}
\AtBeginDocument{%
  }

\usepackage{graphicx}
\usepackage{float}

\usepackage{graphicx}

\usepackage{fancyhdr}

\usepackage{subcaption} 
\usepackage{etoolbox}

\renewcommand\footnotetextcopyrightpermission[1]{}

\begin{document}
\title{One-Shot Pose-Driving  Face Animation Platform}

\pagestyle{empty}
% \author{  
% \resizebox{\linewidth}{!}{
% \begin{tabular}{cccc}
%   He Feng   &Donglin Di*&Wei Chen &Tonghua Su  \\
%   fenghe021209@gmail.com&didonglin@lixiang.com& chenwei10@lixiang.com& thsu@hit.edu.cn\\
%   Harbin Institute of Technology&Space AI, Li Auto&Space AI, Li Auto&Harbin Institute of Technology\\
%   Harbin, China&Beijin, China &Beijin, China &Harbin, China\\
% \end{tabular}}
% }
% \authornote{Donglin Di is the corresponding author.}

\author{He Feng}
\email{fenghe021209@gmail.com}
\affiliation{%
  \institution{Harbin Institute of Technology}
  \city{Harbin}
  \country{China}
}

\author{Donglin Di}
\authornote{Donglin Di is the corresponding author.}
\email{didonglin@lixiang.com}
\affiliation{%
  \institution{Space AI, Li Auto}
  \city{Beijing}
  \country{China}
}

\author{Yongjia Ma}
\email{maguire9993@gmail.com}
\affiliation{%
  \institution{Space AI, Li Auto}
  \city{Beijing}
  \country{China}
}

\author{Wei Chen}
\email{chenwei10@lixiang.com}
\affiliation{%
  \institution{Space AI, Li Auto}
  \city{Beijing}
  \country{China}
}

\author{Tonghua Su}
\email{thsu@hit.edu.cn}
\affiliation{%
  \institution{Harbin Institute of Technology}
  \city{Harbin}
  \country{China}
}

\renewcommand{\shortauthors}{He Feng, Donglin Di, Yongjia Ma, Wei Chen, \& Tonghua Su}

\begin{abstract}
% The goal of face animation is to create dynamic and expressive talking head videos from single  reference face, leveraging driving conditions from video or audio inputs. Existing methods often require fine-tuning for specific identities and fail to produce expressive videos due to limited capability of their Wav2Pose modules.  To achieve one-shot and consecutive talking head video generation, we adopted and refined an existing Image2Video method by  borrowing Face Locator and Motion Frame mechanism from previous study, subsequently fine-tuning the model on large human face video datasets.  Our improvement  enhanced its capability to generate high-quality and expressive talking head videos. We then developed a demo platform leveraging the Gradio framework, facilitating rapid creation of customized talking head videos for users.

The objective of face animation is to generate dynamic and expressive talking head videos from a single reference face, utilizing driving conditions derived from either video or audio inputs. Current approaches often require fine-tuning for specific identities and frequently fail to produce expressive videos due to the limited effectiveness of Wav2Pose modules.
To facilitate the generation of one-shot and more consecutive talking head videos, we refine an existing Image2Video model by integrating a Face Locator and Motion Frame mechanism. We subsequently optimize the model using extensive human face video datasets, significantly enhancing its ability to produce high-quality and expressive talking head videos.
Additionally, we develop a demo platform using the Gradio framework, which streamlines the process, enabling users to quickly create customized talking head videos.
\end{abstract}

% \begin{CCSXML}
% <ccs2012>
%    <concept>
%        <concept_id>10010147.10010178</concept_id>
%        <concept_desc>Computing methodologies~Artificial intelligence</concept_desc>
%        <concept_significance>500</concept_significance>
%        </concept>
%    <concept>
%        <concept_id>10003120.10003145</concept_id>
%        <concept_desc>Human-centered computing~Visualization</concept_desc>
%        <concept_significance>500</concept_significance>
%        </concept>
%  </ccs2012>
% \end{CCSXML}

% \ccsdesc[500]{Computing methodologies~Artificial intelligence}
% \ccsdesc[500]{Human-centered computing~Visualization}

% \keywords{Do, Not, Us, This, Code, Put, the, Correct, Terms, for,
%   Your, Paper}
% \keywords{Posing-driving Face Animation,  Talking Head Video Generation  
%   }

\begin{teaserfigure}
  \centering
  \includegraphics[width=\textwidth]{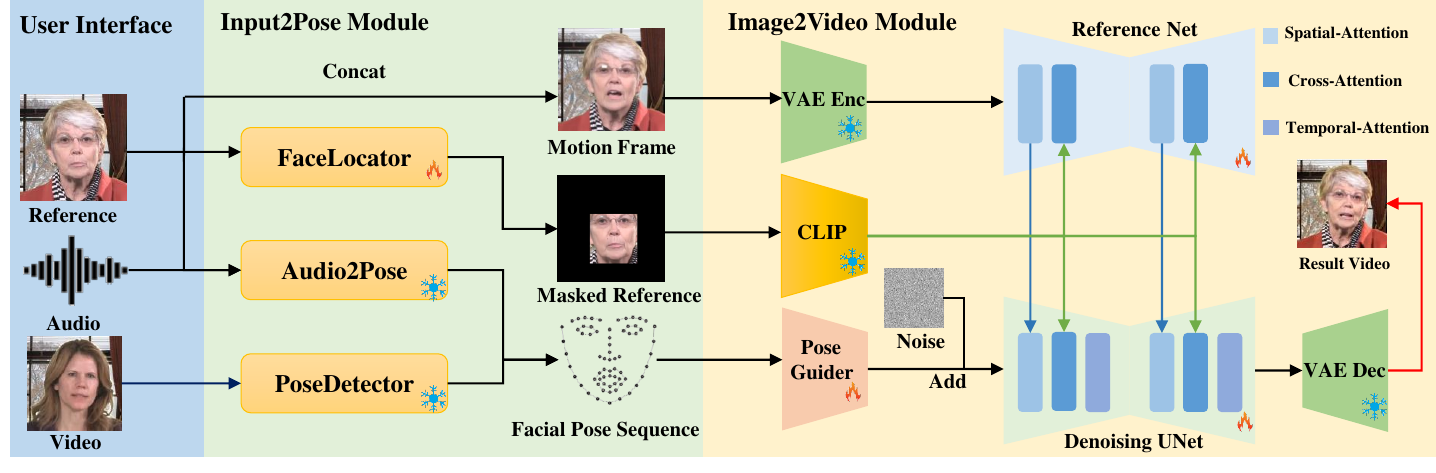}
  \caption{\textbf{Illustration of the face animation platform pipeline.}}
  \Description{Enjoying the baseball game from the third-base
  seats. Ichiro Suzuki preparing to bat.}
  \label{fig:teaser}
\end{teaserfigure}
\maketitle
% \received{20 February 2007}
% \received[revised]{12 March 2009}
% \received[accepted]{5 June 2009}
\section{Introduction }
The core of  face animation lies in accurately capturing a wide range of expressions and movements within a driving pose sequence while preserving  rich facial details of the reference face.
Methods to animate  face can be broadly classified as video-driven, audio-driven, or a blend of both. Video-driven methods \cite{hong23implicit,hong2023dagan,drobyshev2022megaportraits,ni2023cross} extract facial landmarks from driving video, called pose sequences, or extract other control signals. These pose sequences are then used to animate arbitrary reference face images and generate talking head videos that mirror the facial expressions and motions observed in the driving video.  In contrast,  audio-driven methods, such as  \cite{zhang2022sadtalker,ye2023geneface++,shen2023difftalk,prajwal2020lip,ye2024real3d} obtain driving signals or generate 3D intermediate representations from audio input and reference image.  Despite their innovative designs, these methods encounter common challenges: (1) a deficiency in consecutiveness and expressiveness, especially in scenarios involving significant facial movements;   (2) instability of the video background, excluding the subjects;  and (3) the necessity for identity-specific fine-tuning, lack of generalizability.

To address the aforementioned challenges, we employ an Image2Video model \cite{hu2023animate}, further  fine-tuning it on high-quality talking head video datasets, facilitating one-shot talking head video generation. To produce videos with enhanced continuity and minimal background artifacts, we integrate two pioneering modules from \cite{tian2024emo}:  Face Locator and  Motion Frame mechanism.  Face Locator accurately identifies the location for generating the animated face, while  Motion Frame mechanism improves video coherence by randomly selecting and inputting several consecutive frames from the driving video, along with the reference image as input of following network during training.  Additionally, we  developed a user-friendly platform that enables users to upload a reference image and choose from three methods to generate pose sequences for creating personalized talking head videos.

% \begin{itemize}
%     \item 
%     We  introduced two modules into AnimateAnyone, enhancing its capability to produce high-quality, consecutive talking head videos.
%     \item
%     We fine-tuned AnimateAnyone on large human face video datasets, improving its ability to accurately capture facial details and generate  talking head videos that preserve identity of reference face.
%     \item 
%      We developed a user-friendly demo platform that enables any user to create their own customized talking head video anytime and anywhere.

% \end{itemize}

\vspace{-10pt}
\section{Method}
Our method is based on AnimateAnyone, which comprises a Reference Net extracting features from the reference image, a Denoising UNet that denoises multiframe noise and generates videos,  a pre-trained CLIP \cite{radford2021learning} guides the video generation process through cross-attention. The pose sequence, obtained from either DWPose \cite{yang2023effective} or the Audio2Pose module from \cite{wei2024aniportrait}, is input into the Pose Guider, a lightweight CNN, which transforms it into latent space and combines it with noisy latents for facial animation.
Face Locator first locates the facial area from the reference image, then generates a masked reference face to explicitly guide the model in generating facial areas while ensuring the stability of the background. The Motion Frame mechanism operates by randomly selecting consecutive frames during training, which are then channel-wise concatenated with the reference face. This concatenated input is subsequently fed into the Reference Net to extract feature.

The integration of  Face Locator  and Motion Frame mechanism enhances the talking head videos generation capabilities of AnimateAnyone, without notably extending the training and inference time. Our training protocol adheres to the open source code provided by Moore Threads, with CelebV-HQ \cite{zhu2022celebvhq} and HDTF \cite{zhang2021flow} serving as training data sets.

% The \textit{Face Locator} (lightweight MLP) utilizes masks of facial regions extracted from the reference face as its input. Its output is element-wise integrated with the noise input of a \textit{Denoising Unet}, thereby explicitly directing the model to generate facial areas while ensuring the stability of the background. The \textit{Motion Frame} mechanism operates by randomly selecting consecutive frames during training, which are then channel-wise concatenated with the reference face. This concatenated input is subsequently fed into the \textit{Reference Net}. %
% The enhanced AnimateAnyone model processes a facial image and a pose sequence by initially extracting features through the CLIP model and a Reference Network, while the pose sequence undergoes feature extraction via the Pose Guidance system. Features are also derived from a binary facial region map through the Face Locator. These features, integrated with multi-frame noise, guide the synthesis of the talking head video.(random sampling  n frames during training and concatenate them along the channel dimension with the reference image)

\begin{figure}[t]
    \centering
    \includegraphics[width=\linewidth,height=5cm]{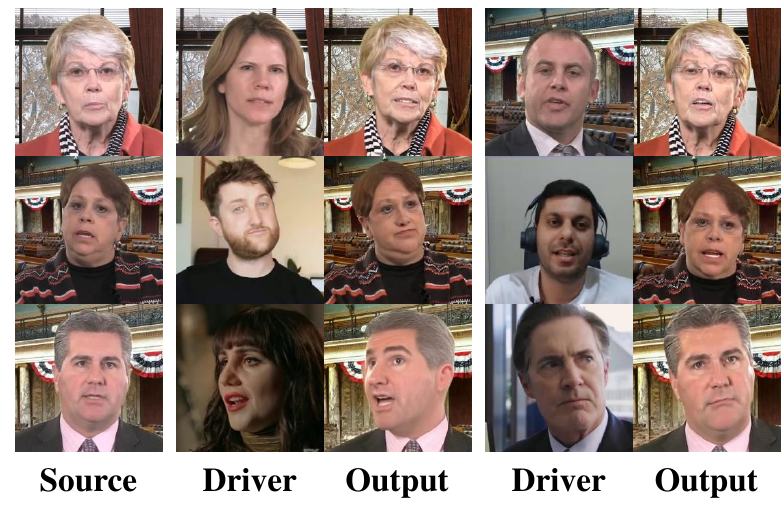}
    \caption {\textbf{Selected running results of our platform. }}
    \label{fig:enter-label}
\end{figure}

% \begin{figure}[H]
%     \centering
%     \includegraphics[width=\linewidth,height=5cm]{test_crop (1).pdf}
%     \caption {\textbf{Selected running results of our platform. }}
%     \label{fig:enter-label}
% \end{figure}
\vspace*{-10pt}

\section{Platform}
Figure \ref{fig:teaser} shows the workflow of our platform.
Our platform can be divided into two main modules: the Input2Pose module and the Image2Video module.

\textbf{Input2Pose module. }In this module, users are initially required to upload a reference face image. Subsequently, they are to select one of the three methods (selecting from our offerings, extracting from uploaded videos, or predicting from uploaded audio) to acquire a pose sequence. Figure \ref{fig:enter-label} shows some selected running results of our platform.
% It should be noted that a significant difference in facial proportions between the reference image and the pose sequence can adversely affect the output video quality. To address this issue, we applied a cropping pre-process to the reference face, ensuring the quality of the final video.

\textbf{Image2Video module. }After acquiring a pose sequence and reference image, users can generate result video by animating the reference image using the pose sequence. All generated videos are set to a default resolution of 512x512 at 24 frames per second. Users may specify the duration of the generated video based on the length of the pose sequence.

\textbf{Application. }The proposed platform exhibits significant untapped potential in various areas, including personal assistance, intelligent customer service, and digital education. Within the sphere of digital education, it can be used to produce digital teachers, boosting student engagement and reducing teacher workload.

\section{Conclusion}
In this work, we introduce a platform for one-shot pose-driving human face animation.   Our platform facilitates the generation of high-quality and expressive talking head videos from a single reference facial image, eliminating the requirement for fine-tuning for specific identities. The platform offers users convenient, personalized talking head video generation services, bridging the gap between related technological research and user interaction. Our current method employs a dual U-Net architecture and still has room for optimization in inference and training speeds to ensure  the capability for real-time animation.  Future research should prioritize enhancing computational efficiency,  for instance, through the adoption of more efficient architectures.

%%
%% The acknowledgments section is defined using the "acks" environment
%% (and NOT an unnumbered section). This ensures the proper
%% identification of the section in the article metadata, and the
%% consistent spelling of the heading.
% \begin{acks}
% To Robert, for the bagels and explaining CMYK and color spaces.
% \end{acks}

\bibliographystyle{ACM-Reference-Format}
\bibliography{ref}

%%% -*-BibTeX-*-
%%% Do NOT edit. File created by BibTeX with style
%%% ACM-Reference-Format-Journals [18-Jan-2012].

\begin{thebibliography}{16}

%%% ====================================================================
%%% NOTE TO THE USER: you can override these defaults by providing
%%% customized versions of any of these macros before the \bibliography
%%% command.  Each of them MUST provide its own final punctuation,
%%% except for \shownote{}, \showDOI{}, and \showURL{}.  The latter two
%%% do not use final punctuation, in order to avoid confusing it with
%%% the Web address.
%%%
%%% To suppress output of a particular field, define its macro to expand
%%% to an empty string, or better, \unskip, like this:
%%%
%%% \newcommand{\showDOI}[1]{\unskip}   % LaTeX syntax
%%%
%%% \def \showDOI #1{\unskip}           % plain TeX syntax
%%%
%%% ====================================================================

\ifx \showCODEN    \undefined \def \showCODEN     #1{\unskip}     \fi
\ifx \showDOI      \undefined \def \showDOI       #1{#1}\fi
\ifx \showISBNx    \undefined \def \showISBNx     #1{\unskip}     \fi
\ifx \showISBNxiii \undefined \def \showISBNxiii  #1{\unskip}     \fi
\ifx \showISSN     \undefined \def \showISSN      #1{\unskip}     \fi
\ifx \showLCCN     \undefined \def \showLCCN      #1{\unskip}     \fi
\ifx \shownote     \undefined \def \shownote      #1{#1}          \fi
\ifx \showarticletitle \undefined \def \showarticletitle #1{#1}   \fi
\ifx \showURL      \undefined \def \showURL       {\relax}        \fi
% The following commands are used for tagged output and should be
% invisible to TeX
\providecommand\bibfield[2]{#2}
\providecommand\bibinfo[2]{#2}
\providecommand\natexlab[1]{#1}
\providecommand\showeprint[2][]{arXiv:#2}

\bibitem[Drobyshev et~al\mbox{.}(2022)]%
        {drobyshev2022megaportraits}
\bibfield{author}{\bibinfo{person}{Nikita Drobyshev}, \bibinfo{person}{Jenya Chelishev}, \bibinfo{person}{Taras Khakhulin}, \bibinfo{person}{Aleksei Ivakhnenko}, \bibinfo{person}{Victor Lempitsky}, {and} \bibinfo{person}{Egor Zakharov}.} \bibinfo{year}{2022}\natexlab{}.
\newblock \showarticletitle{Megaportraits: One-shot megapixel neural head avatars}. In \bibinfo{booktitle}{\emph{Proceedings of the 30th ACM International Conference on Multimedia}}. \bibinfo{pages}{2663--2671}.
\newblock


\bibitem[Hong et~al\mbox{.}(2023)]%
        {hong2023dagan}
\bibfield{author}{\bibinfo{person}{Fa-Ting Hong}, \bibinfo{person}{}, \bibinfo{person}{Li Shen}, {and} \bibinfo{person}{Dan Xu}.} \bibinfo{year}{2023}\natexlab{}.
\newblock \showarticletitle{DaGAN++: Depth-Aware Generative Adversarial Network for Talking Head Video Generation}.
\newblock \bibinfo{journal}{\emph{IEEE Transactions on Pattern Analysis and Machine Intelligence (TPAMI)}} (\bibinfo{year}{2023}).
\newblock


\bibitem[Hong and Xu(2023)]%
        {hong23implicit}
\bibfield{author}{\bibinfo{person}{Fa-Ting Hong} {and} \bibinfo{person}{Dan Xu}.} \bibinfo{year}{2023}\natexlab{}.
\newblock \showarticletitle{Implicit Identity Representation Conditioned Memory Compensation Network for Talking Head video Generation}. In \bibinfo{booktitle}{\emph{ICCV}}.
\newblock


\bibitem[Hu et~al\mbox{.}(2023)]%
        {hu2023animate}
\bibfield{author}{\bibinfo{person}{Li Hu}, \bibinfo{person}{Xin Gao}, \bibinfo{person}{Peng Zhang}, \bibinfo{person}{Ke Sun}, \bibinfo{person}{Bang Zhang}, {and} \bibinfo{person}{Liefeng Bo}.} \bibinfo{year}{2023}\natexlab{}.
\newblock \showarticletitle{Animate anyone: Consistent and controllable image-to-video synthesis for character animation}.
\newblock \bibinfo{journal}{\emph{arXiv preprint arXiv:2311.17117}} (\bibinfo{year}{2023}).
\newblock


\bibitem[Ni et~al\mbox{.}(2023)]%
        {ni2023cross}
\bibfield{author}{\bibinfo{person}{Haomiao Ni}, \bibinfo{person}{Yihao Liu}, \bibinfo{person}{Sharon~X Huang}, {and} \bibinfo{person}{Yuan Xue}.} \bibinfo{year}{2023}\natexlab{}.
\newblock \showarticletitle{Cross-identity video motion retargeting with joint transformation and synthesis}. In \bibinfo{booktitle}{\emph{Proceedings of the IEEE/CVF Winter Conference on Applications of Computer Vision}}. \bibinfo{pages}{412--422}.
\newblock


\bibitem[Prajwal et~al\mbox{.}(2020)]%
        {prajwal2020lip}
\bibfield{author}{\bibinfo{person}{KR Prajwal}, \bibinfo{person}{Rudrabha Mukhopadhyay}, \bibinfo{person}{Vinay~P Namboodiri}, {and} \bibinfo{person}{CV Jawahar}.} \bibinfo{year}{2020}\natexlab{}.
\newblock \showarticletitle{A lip sync expert is all you need for speech to lip generation in the wild}. In \bibinfo{booktitle}{\emph{Proceedings of the 28th ACM international conference on multimedia}}. \bibinfo{pages}{484--492}.
\newblock


\bibitem[Radford et~al\mbox{.}(2021)]%
        {radford2021learning}
\bibfield{author}{\bibinfo{person}{Alec Radford}, \bibinfo{person}{Jong~Wook Kim}, \bibinfo{person}{Chris Hallacy}, \bibinfo{person}{Aditya Ramesh}, \bibinfo{person}{Gabriel Goh}, \bibinfo{person}{Sandhini Agarwal}, \bibinfo{person}{Girish Sastry}, \bibinfo{person}{Amanda Askell}, \bibinfo{person}{Pamela Mishkin}, \bibinfo{person}{Jack Clark}, {et~al\mbox{.}}} \bibinfo{year}{2021}\natexlab{}.
\newblock \showarticletitle{Learning transferable visual models from natural language supervision}. In \bibinfo{booktitle}{\emph{International conference on machine learning}}. PMLR, \bibinfo{pages}{8748--8763}.
\newblock


\bibitem[Shen et~al\mbox{.}(2023)]%
        {shen2023difftalk}
\bibfield{author}{\bibinfo{person}{Shuai Shen}, \bibinfo{person}{Wenliang Zhao}, \bibinfo{person}{Zibin Meng}, \bibinfo{person}{Wanhua Li}, \bibinfo{person}{Zheng Zhu}, \bibinfo{person}{Jie Zhou}, {and} \bibinfo{person}{Jiwen Lu}.} \bibinfo{year}{2023}\natexlab{}.
\newblock \showarticletitle{Difftalk: Crafting diffusion models for generalized audio-driven portraits animation}. In \bibinfo{booktitle}{\emph{Proceedings of the IEEE/CVF Conference on Computer Vision and Pattern Recognition}}. \bibinfo{pages}{1982--1991}.
\newblock


\bibitem[Tian et~al\mbox{.}(2024)]%
        {tian2024emo}
\bibfield{author}{\bibinfo{person}{Linrui Tian}, \bibinfo{person}{Qi Wang}, \bibinfo{person}{Bang Zhang}, {and} \bibinfo{person}{Liefeng Bo}.} \bibinfo{year}{2024}\natexlab{}.
\newblock \showarticletitle{EMO: Emote Portrait Alive-Generating Expressive Portrait Videos with Audio2Video Diffusion Model under Weak Conditions}.
\newblock \bibinfo{journal}{\emph{arXiv preprint arXiv:2402.17485}} (\bibinfo{year}{2024}).
\newblock


\bibitem[Wei et~al\mbox{.}(2024)]%
        {wei2024aniportrait}
\bibfield{author}{\bibinfo{person}{Huawei Wei}, \bibinfo{person}{Zejun Yang}, {and} \bibinfo{person}{Zhisheng Wang}.} \bibinfo{year}{2024}\natexlab{}.
\newblock \showarticletitle{Aniportrait: Audio-driven synthesis of photorealistic portrait animation}.
\newblock \bibinfo{journal}{\emph{arXiv preprint arXiv:2403.17694}} (\bibinfo{year}{2024}).
\newblock


\bibitem[Yang et~al\mbox{.}(2023)]%
        {yang2023effective}
\bibfield{author}{\bibinfo{person}{Zhendong Yang}, \bibinfo{person}{Ailing Zeng}, \bibinfo{person}{Chun Yuan}, {and} \bibinfo{person}{Yu Li}.} \bibinfo{year}{2023}\natexlab{}.
\newblock \showarticletitle{Effective whole-body pose estimation with two-stages distillation}. In \bibinfo{booktitle}{\emph{Proceedings of the IEEE/CVF International Conference on Computer Vision}}. \bibinfo{pages}{4210--4220}.
\newblock


\bibitem[Ye et~al\mbox{.}(2023)]%
        {ye2023geneface++}
\bibfield{author}{\bibinfo{person}{Zhenhui Ye}, \bibinfo{person}{Jinzheng He}, \bibinfo{person}{Ziyue Jiang}, \bibinfo{person}{Rongjie Huang}, \bibinfo{person}{Jiawei Huang}, \bibinfo{person}{Jinglin Liu}, \bibinfo{person}{Yi Ren}, \bibinfo{person}{Xiang Yin}, \bibinfo{person}{Zejun Ma}, {and} \bibinfo{person}{Zhou Zhao}.} \bibinfo{year}{2023}\natexlab{}.
\newblock \showarticletitle{GeneFace++: Generalized and Stable Real-Time Audio-Driven 3D Talking Face Generation}.
\newblock \bibinfo{journal}{\emph{arXiv preprint arXiv:2305.00787}} (\bibinfo{year}{2023}).
\newblock


\bibitem[Ye et~al\mbox{.}(2024)]%
        {ye2024real3d}
\bibfield{author}{\bibinfo{person}{Zhenhui Ye}, \bibinfo{person}{Tianyun Zhong}, \bibinfo{person}{Yi Ren}, \bibinfo{person}{Jiaqi Yang}, \bibinfo{person}{Weichuang Li}, \bibinfo{person}{Jiawei Huang}, \bibinfo{person}{Ziyue Jiang}, \bibinfo{person}{Jinzheng He}, \bibinfo{person}{Rongjie Huang}, \bibinfo{person}{Jinglin Liu}, {et~al\mbox{.}}} \bibinfo{year}{2024}\natexlab{}.
\newblock \showarticletitle{Real3d-portrait: One-shot realistic 3d talking portrait synthesis}.
\newblock \bibinfo{journal}{\emph{arXiv preprint arXiv:2401.08503}} (\bibinfo{year}{2024}).
\newblock


\bibitem[Zhang et~al\mbox{.}(2022)]%
        {zhang2022sadtalker}
\bibfield{author}{\bibinfo{person}{Wenxuan Zhang}, \bibinfo{person}{Xiaodong Cun}, \bibinfo{person}{Xuan Wang}, \bibinfo{person}{Yong Zhang}, \bibinfo{person}{Xi Shen}, \bibinfo{person}{Yu Guo}, \bibinfo{person}{Ying Shan}, {and} \bibinfo{person}{Fei Wang}.} \bibinfo{year}{2022}\natexlab{}.
\newblock \showarticletitle{SadTalker: Learning Realistic 3D Motion Coefficients for Stylized Audio-Driven Single Image Talking Face Animation}.
\newblock \bibinfo{journal}{\emph{arXiv preprint arXiv:2211.12194}} (\bibinfo{year}{2022}).
\newblock


\bibitem[Zhang et~al\mbox{.}(2021)]%
        {zhang2021flow}
\bibfield{author}{\bibinfo{person}{Zhimeng Zhang}, \bibinfo{person}{Lincheng Li}, \bibinfo{person}{Yu Ding}, {and} \bibinfo{person}{Changjie Fan}.} \bibinfo{year}{2021}\natexlab{}.
\newblock \showarticletitle{Flow-guided one-shot talking face generation with a high-resolution audio-visual dataset}. In \bibinfo{booktitle}{\emph{Proceedings of the IEEE/CVF Conference on Computer Vision and Pattern Recognition}}. \bibinfo{pages}{3661--3670}.
\newblock


\bibitem[Zhu et~al\mbox{.}(2022)]%
        {zhu2022celebvhq}
\bibfield{author}{\bibinfo{person}{Hao Zhu}, \bibinfo{person}{Wayne Wu}, \bibinfo{person}{Wentao Zhu}, \bibinfo{person}{Liming Jiang}, \bibinfo{person}{Siwei Tang}, \bibinfo{person}{Li Zhang}, \bibinfo{person}{Ziwei Liu}, {and} \bibinfo{person}{Chen~Change Loy}.} \bibinfo{year}{2022}\natexlab{}.
\newblock \showarticletitle{{CelebV-HQ}: A Large-Scale Video Facial Attributes Dataset}. In \bibinfo{booktitle}{\emph{ECCV}}.
\newblock


\end{thebibliography}

\end{document}